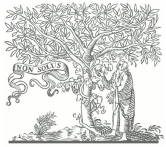

# Visual appearance based person retrieval in unconstrained environment videos☆

Hiren Galiyawala [a,][*], Mehul S. Raval [a], Shivansh Dave [b]

[a] *Department of Information and Communication Technology, School of Technology, Pandit Deendayal Petroleum University, India*
[b] *Department of Computer Science, University of Windsor, Canada*

ABSTRACT

Visual appearance-based person retrieval is a challenging problem in surveillance. It uses attributes like height, cloth color, cloth type and gender to describe a human. Such attributes are known as soft biometrics. This paper proposes person retrieval from surveillance video using height, torso cloth type, torso cloth color and gender. The approach introduces an adaptive torso patch extraction and bounding box regression to improve the retrieval. The algorithm uses fine-tuned Mask R-CNN and DenseNet-169 for person detection and attribute classification respectively. The performance is analyzed on AVSS 2018 challenge II dataset and it achieves 11.35% improvement over state-of-the-art based on average Intersection over Union measure.

© 2019.

## 1. Introduction

Modern society seeks a right and secure place for living. Many countries have surveillance systems for securing public places like shopping malls, cross-sections, parks, railway stations, and government offices. Surveillance systems are useful in many applications like crime investigation, traffic monitoring, and person localization. Traditional biometric (e.g., face) fails to locate a person in the surveillance video due to low resolution and unconstrained environment. A typical surveillance video frame is shown in Fig. 1. Many attributes of the person like cloth color, gender, and cloth type are visible in the frame and they are known as soft biometrics [1-4]. A visual appearance-based description uses soft biometrics like height, gender, cloth color, torso type, and leg type. For example, person in the bounding box of Fig. 1 is described as *a female with brown t-shirt and blue shorts*.

A manual scouring of feeds for person retrieval is becoming difficult due to volume, velocity, and veracity of the videos. Thus, automatic person retrieval is gaining attention of the research community. The research work [6-8] uses handcrafted features to classify the soft biometrics. Halstead et al. [6] propose an approach that builds a query using torso and leg color, height, build and leg clothing type. Particle filter then search through video sequences. The work uses fix body-region to search torso and leg color. The extracted body region includes background pixels leading to incorrect color classification. Denman et al. [7] create an avatar using a search query in the form of channel representation. The search query uses height, dominant color (torso and leg), and clothing type (torso and leg). The approach by the Martinho-Corbishley et al. [8] generates hand-crafted features based on Ensemble of Localised Features (ELF) descriptor and uses Extra Tree Classification (ETC) algorithm for identification. The paper introduces Soft Biometric Retrieval (SoBiR) dataset with 8 camera views, 100 persons and categorical annotations. Shah et al. [9] use cloth color and cloth type for person retrieval. Motion segmentation based on frame differencing is used to locate the person(s). The approach cannot detect a stationary person and moving object generates noise during detection. Thus, handcrafted features based approaches may limit the retrieval accuracy.

Deep learning based methodologies are widely used for semantic description based person retrieval due to its efficient learning. Convolutional Neural Network (CNN) based algorithms are reported for person attribute recognition [10-12] as well as for person retrieval [5, 13-16]. Semantic Retrieval Convolution Neural Network (SRCNN) [10] retrieves the soft biometrics with recognition rate of 20.1% and 46.4% at rank-1 for one-shot and multi-shot identification respectively on SoBiR [8] dataset. Li et al. [12] explore the spatial correlations of person attributes with human body structure e.g., hair and glasses are most correlated with the head. They propose Pose Guided Deep Model (PGDM) consisting of coarse pose estimation, body part localization, and fusion of body part features. Pose annotations in the dataset are avoided by PGDM which transfer the existing pose estimation model knowledge to pedestrian attribute database. The evaluation uses RAP [17], PETA [18] and PA-100K [19] large scale pedestrian attribute datasets. Given the input image, Part-based Convolutional Baseline (PCB) [13] produces convolutional descriptor with part-level features. It gives uniform partition on the image without explicitly partitioning it. Refined Part Pooling (RPP) follows uniform partitioning which adaptively improves the partition. It achieves (77.4 ± 4.2)% mAP and (92.3 ± 1.5)% rank-1 accuracy on Market-

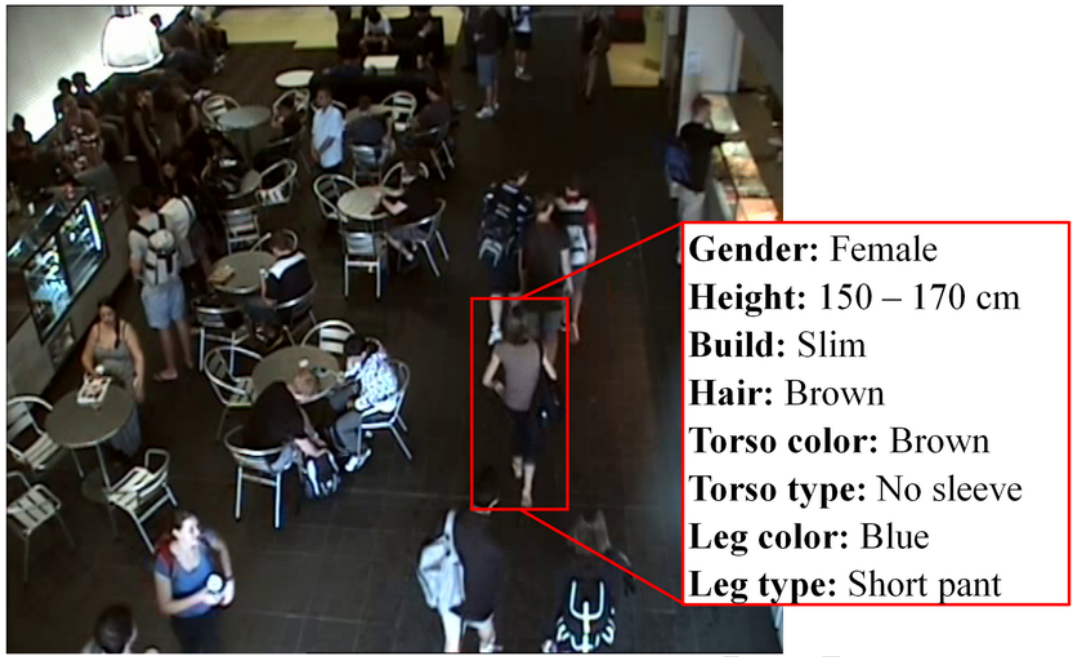

**Fig. 1.** Typical surveillance video frame [5]: Frame in which faces are unrecognizable, persons appear with different poses and occluded.

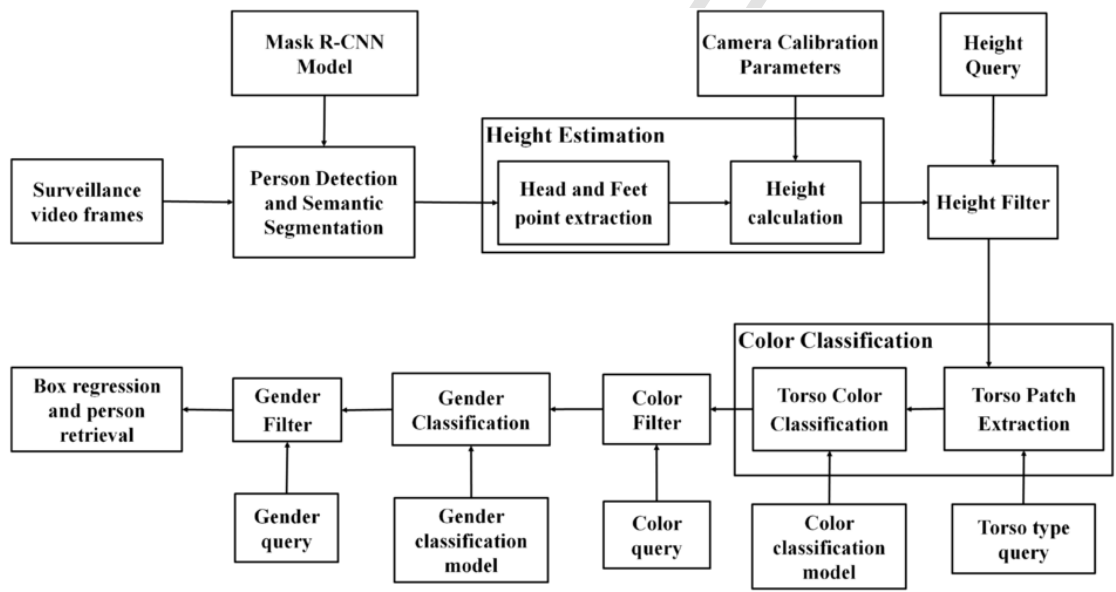

**Fig. 2.** The proposed linear filtering approach of person retrieval.

1501 [20] dataset. They also demonstrate evaluation on DukeMTMC-reID [21] and CUHK03 [22] datasets.

Researchers are extensively using PETA [18], Market-1501 [20], DukeMTMC-reID [21], CUHK03 [22] and RAP [17] large scale pedestrian datasets for person identification or re-identification. These datasets consist of a gallery of cropped images of persons and annotated soft biometric attributes. However, the datasets lack availability of continuous video frames and their semantic annotations are also limited.

Halstead et al. proposed a challenge (AVSS 2018 challenge II) [5] to rectify the above limitations. The challenge is expected to retrieve a person in surveillance video using a semantic query. The outcome of the AVSS challenge II showcased that CNN based algorithms achieved higher accuracy. Galiyawala et al. [15] propose a linear filtering approach which reduces the search space during the retrieval. Semantic segmentation nullifies limitations of the bounding box based approaches. Mask R-CNN [23] is used for person detection and semantic segmentation. The height is estimated using [24] while the other two attributes i.e., torso color and gender are classified using AlexNet [25]. The algorithm achieves an average Intersection-over-Union (IoU) of 0.36 and 52.2% of frames have an IoU greater than 0.4. The baseline approach [7] and [15] uses subset of available soft biometrics, limiting their accuracy. The color patch extraction is also noisy in [15]. Moreover, [15] uses AlexNet [25] which has lower retrieval accuracy compared to deeper networks like ResNet [26] and DenseNet [27].

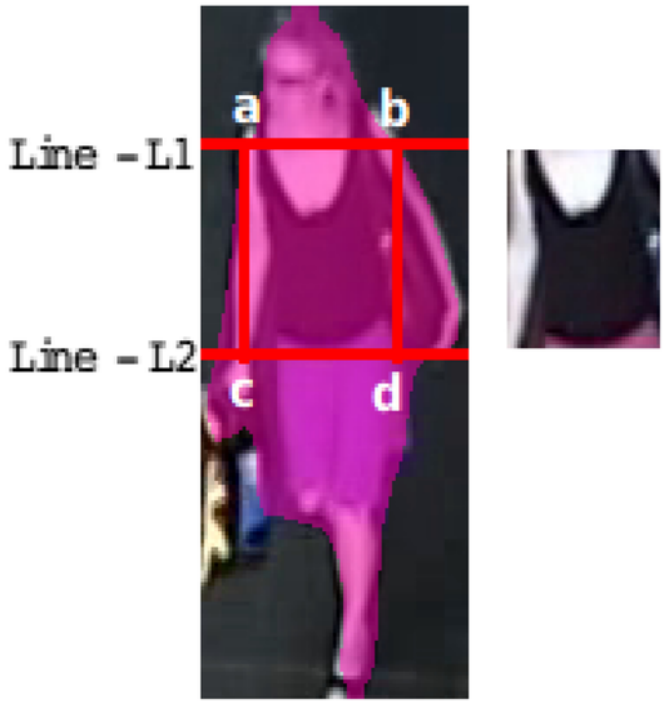

**Fig. 3.** Torso patch extraction.

The transfer learning based approach by Yaguchi et al. [14] also uses Mask R-CNN for person detection and DenseNet-161 for soft biometric classification. The Hamming loss is used to calculate matching score of the semantic query. The retrieved subject with the smallest loss is the target. Schumann et al. [16] use Single-Shot Multibox Detector (SSD) for person detection and refine it using degree of motion in the SSD detected area by background modeling. Such detection methods fail to locate non-moving persons. Four backbone architectures (MobileNet-v1, ResNet-50-v1, DenseNet-121, and DenseNet-169) are used during ensemble classification and predictions are finally fused for final result. The paper uses one more model for pose estimation while extracting the patch for color and texture. In the AVSS challenge II, [14] achieved state-of-the-art performance with an average IoU of 0.51 and [16] had 75.9% of frames with an IoU greater than 0.4. The use of tracking in the video sequences provides superior detection [16].

The paper proposes a person retrieval using height, torso cloth type, torso cloth color, and gender. The algorithm builds on the linear filtering approach [15] for person retrieval. The height is estimated using camera calibration approach [24]. The torso color and gender are identified using CNN. The proposed approach overcomes limitations of [14-16] by; 1. adaptively extracting torso patch; 2. using IoU based bounding box regression; 3. using fewer soft biometrics.

Major contributions of the work are summarized as follows:

(1) The retrieval is improved by adaptively extracting torso patch based on the type given in the query. It reduces noisy pixels and increases classification accuracy compared to static patch extraction [15].
(2) An IoU based regression is applied to predict a bounding box in the frame where person detection using soft biometric fails. The bounding box regressor increases the number of frames with detection and helps to locate the person during partial occlusion, poor illumination, and merging of foreground with background.
(3) The method uses fewer soft biometrics in comparison to [14, 16].

Further, the paper is organized as follows. Section 2 describes the approach with implementation details for color and gender classification. Section 3 discusses experimental results and comparison with previous approaches. Section 4 concludes the paper.

## 2. Proposed approach

This section discusses the semantic person retrieval in surveillance using height, torso type, torso color, and gender. The person's height is estimated using Tsai camera calibration approach [24]. Torso color and gender are classified by fine-tuning DenseNet-169 [27]. The cascaded classifiers for height, torso color, and gender aim to remove non-matching persons at every stage. It reduces the search space of the match. Fig. 2 depicts the algorithmic flow diagram. Mask R-CNN [23] is used as person detector in every surveillance frame. It outputs detected persons with their bounding box, semantic segmentation, and instance segmentation. Semantic segmentation produces close contour (mask) of the person in the bounding box. An instance segmentation distinctly labels detected person(s).

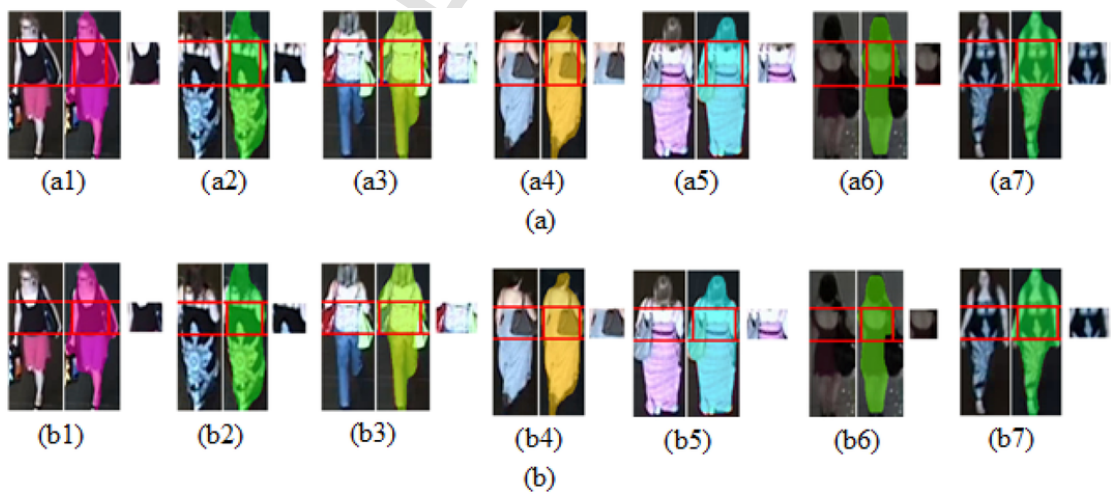

**Fig. 4.** Samples of torso patch extraction: (a) torso patch extraction using static ratio and (b) torso patch extracted using adaptive ratio. In each sample, images from left to right are as follows: the bounding box output of Mask R-CNN, semantic segmentation with patch extraction region and extracted patch.

4    Image and Vision Computing xxx (xxxx) xxx-xxx

**Table 1**
Adaptive ratio according to torso type.

| Clothing type | | Ratio (%) from top of bounding box |
|---|---|---|
| **Torso** | Long sleeve | 20%–48% |
| | short sleeve | 20%–48% |
| | No sleeve | 25%–48% |
| | Indian kurta/dress | 20%–56% |
| **Leg** | Long pants | 56%–84% |
| | Dress | 56%–84% |
| | Skirt | 52%–64% |
| | Long shorts | 56%–68% |
| | Short shorts | 52%–62% |
| | Indian kurta/dress | 75%–90% |

**Table 2**
Soft biometric attributes.

| Attribute | Class |
|---|---|
| Color | Unknown, black, blue, brown, green, grey, orange, pink, purple, red, white, yellow, skin. |
| Torso type | Unknown, long sleeve, short sleeve, no sleeve. |
| Gender | Unknown, male, female. |
| Height | Unknown, very short (130–160 cm), short (150–170 cm), average (160–180 cm), tall (170–190 cm), very tall (180–210 cm). |

The mask is used to extract head and feet points (i.e., image coordinates) for every detected person. These image coordinates are translated into real-world coordinates using [24]. AVSS 2018 challenge II [5] dataset provides 6 calibrated cameras. The real world height is estimated using camera calibration parameters. The algorithm adapts height estimation using camera calibration described in [15]. The estimated height is compared with the height corresponding to class (e.g., *medium*) provided with the query. The module outputs person(s) which matches the query height class. Thus, height acts as an initial filter to reduce the search space in the frame. The output may contain many persons; due to multiple matches within the same height class.

Based on the torso type in the query, region for torso patch extraction is adapted for all detections. The torso color is classified using DenseNet-169 trained on ImageNet [28] dataset and it is fine-tuned on custom color patches using transfer learning. AVSS 2018 challenge II [5] dataset provides two torso colors namely; color-1 and color-2. The algorithm filters the persons for color-1. In case a match is not found then torso color-2 is used for person retrieval (rank-2 match).

In case of multiple retrievals, further filtering is done using gender classification. Gender is classified using DenseNet-169 trained on ImageNet [28] dataset and it is fine-tuned on person images collected from standard datasets [17, 18, 21]. An IoU based bounding box regressor improves the detection. It predicts the box in the frame where person detection fails for all three soft biometrics. Further subsections discuss torso patch extraction, torso color, and gender classification, and IoU based box regression for person retrieval.

*2.1. Cloth color detection and color filtering*

The color classification block extracts the torso patch and assigns color label to it. Fig. 3 shows the general strategy for torso patch extraction. The region covered by segmentation mask, line - L1, and line - L2 is the required patch. Semantic segmentation allows the background free patch extraction [15]. The locations of L1 and L2 are decided by torso type given in the query. It helps to extract torso patches without unwanted pixels from the human body covered by region enclosed by *a, b, c, d* in the Fig. 3.

Irrespective of the clothing type, the approach in [15] uses static ratio for patch extraction. Fig. 4 (a) shows static patch extraction where L1 is fixed at 20% and L2 at 50% from the top of the bounding box. Thus, it includes unwanted pixels in the bounding box, making classification noisy. Extracting the torso patch according to the clothing type resolves the problem. Table 1 shows the dynamic range for patch extraction according to torso type and leg type clothing.

Fig. 4 shows patch extraction samples for the torso type *no sleeve*. The algorithm does not use leg color in the retrieval process,

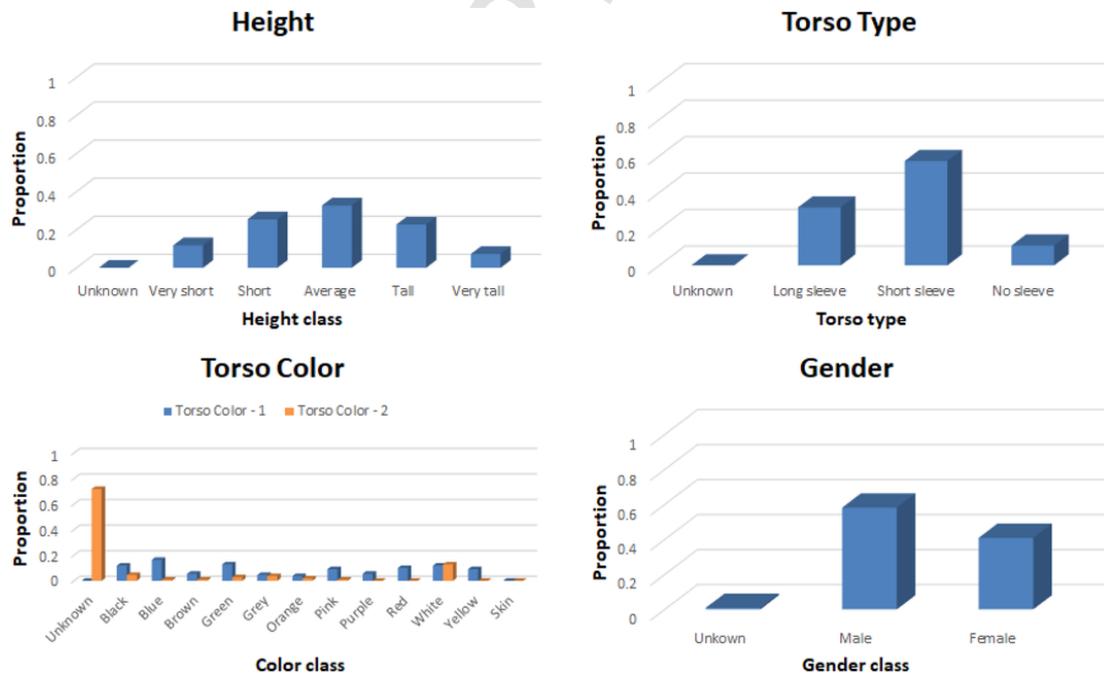

**Fig. 5.** Distribution of class for height, torso type, torso color-1, torso color-2 and gender in training dataset.

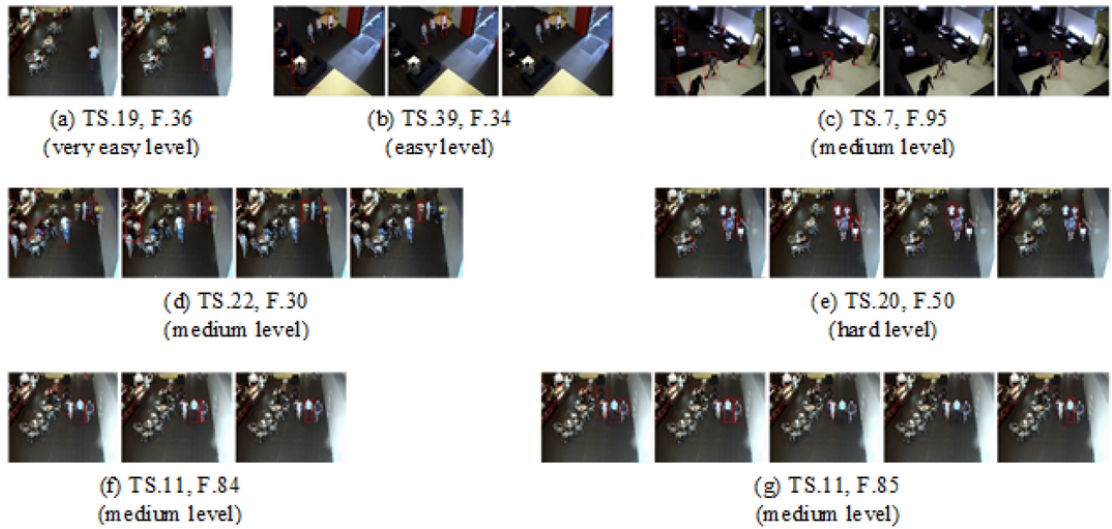

**Fig. 6.** Samples of correctly retrieved person using semantic description: abbreviation, e.g., TS.02, F.33 (very easy level) indicates Test Sequence 02 with frame number 33 and difficulty level, very easy in the test data set. Images from left to right are Mask R-CNN person detection, height filtering, cloth color filtering, gender filtering and output of IoU based regression.

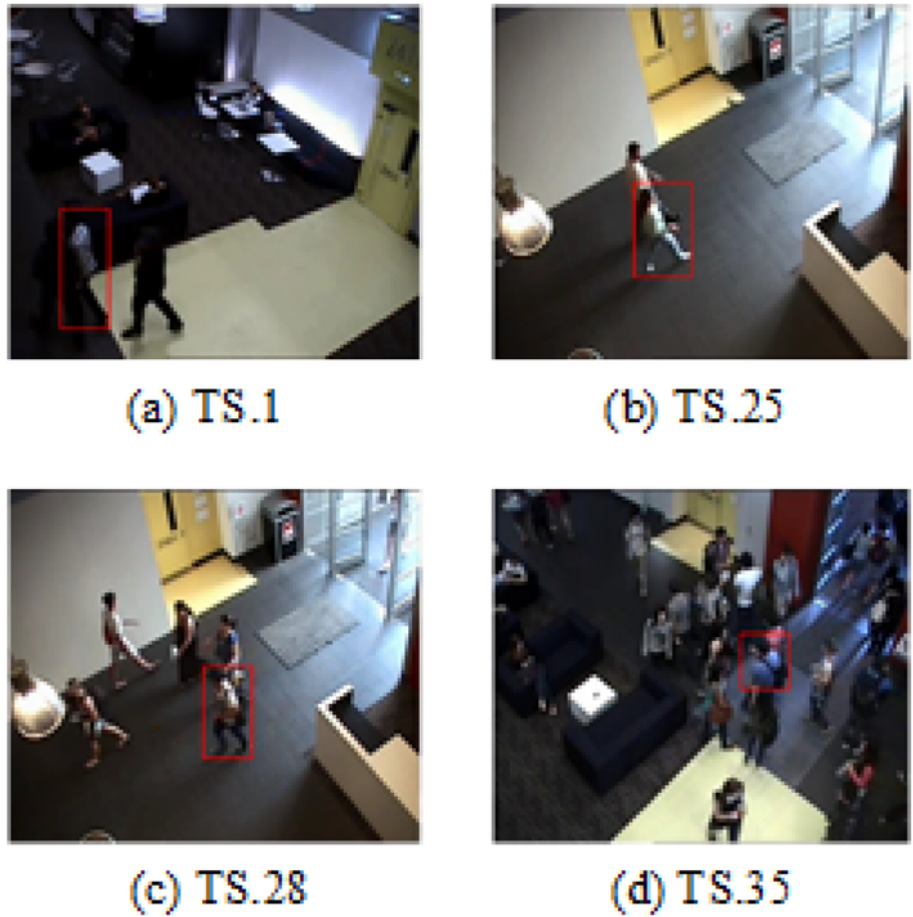

**Fig. 7.** Sequences where retrieval failed.

hence only the torso patch extraction samples are shown in Fig. 4 (b) using the dynamic ratio (Table 1). It can be observed in Fig. 4 (a1) that the extracted patch includes many skin pixels while corresponding patch (Fig. 4 (b1)) has very few skin pixels. It reduces chances of color misclassification. Similarly, Fig. 4 (a2) and (b2) shows back pose of the person. Patch extracted with static ratio includes skin as well as hair pixels. With adaptive ratio the patch does not include those pix-els. Thus, adaptive patch extraction using clothing type improves the torso color classification accuracy.

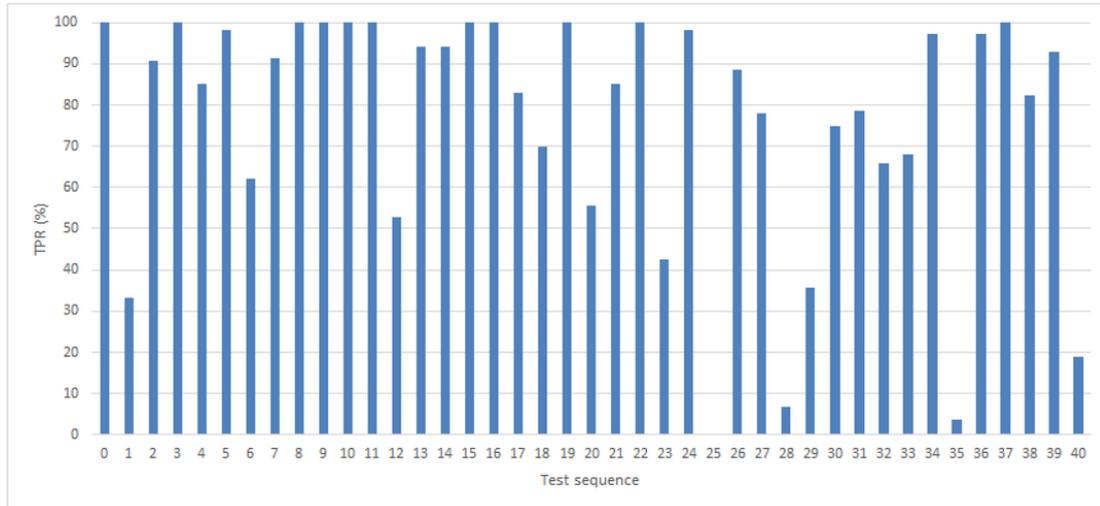

**Fig. 8.** TPR(%) for each test sequences.

**Table 3**
Performance comparison with state-of-the-art methods on AVSS 2018 challenge II (Task-2) dataset [5].

| Methods | Average IoU | $IoU \geq 0.4$ |
| --- | --- | --- |
| Baseline [7] | 0.290 | 0493 |
| [15] | 0.363 | 0.522 |
| [16] | 0.503 | **0.759** |
| [14] | 0.511 | 0.669 |
| Proposed | **0.569** | 0.746 |

The extracted patch is classified using fine-tuned DenseNet-169. The assorted color for each patch is compared with the query color and cleared by the color filter. The filter may produce many persons people with similar torso color. Thus, persons with similar height and torso color are available for further filtering.

### 2.2. Gender classification

Gender block classifies each person filtered through height and color filter. Gender is classified using DenseNet-169 which is fine-tuned using full body images. The classified result is matched with the query. In case of multiple match, further selection is based on probabilities of the final SoftMax classifier.

### 2.3. IoU based box regression

An IoU based regression predicts bounding box (containing the match) for a frame in which detection with soft biometrics fails.

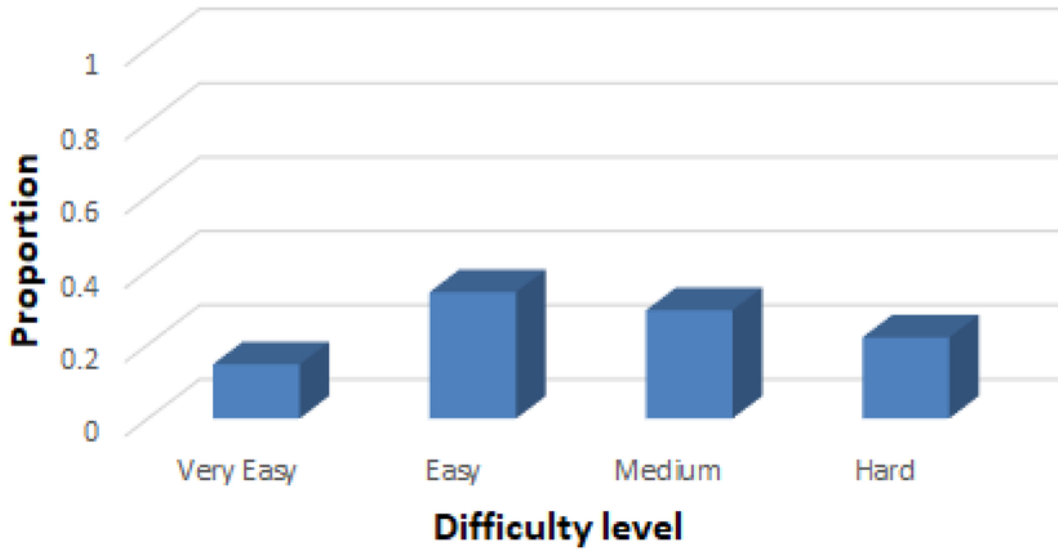

**Fig. 9.** Difficulty level wise distribution of test sequences.

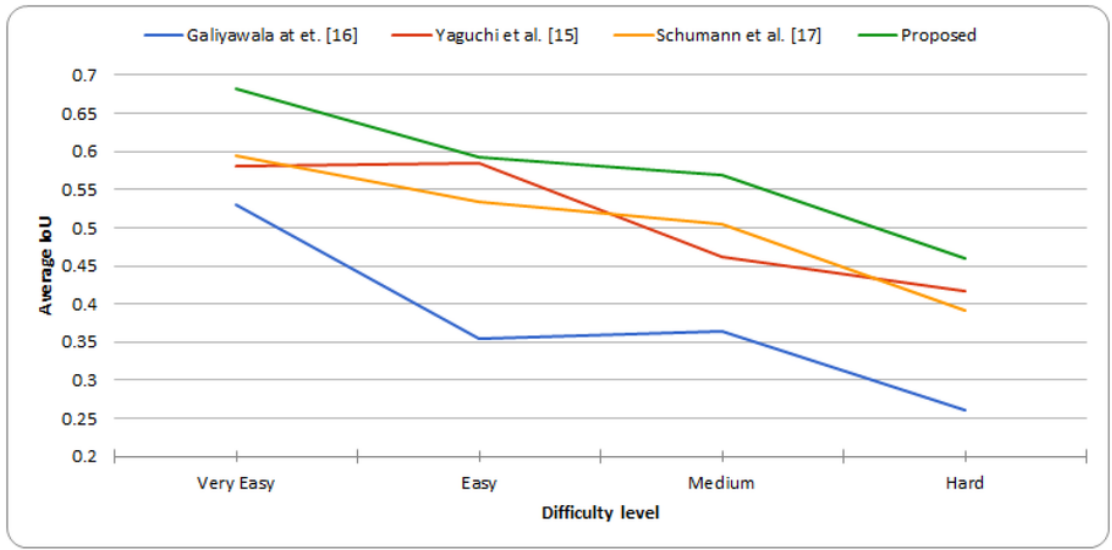

(a)

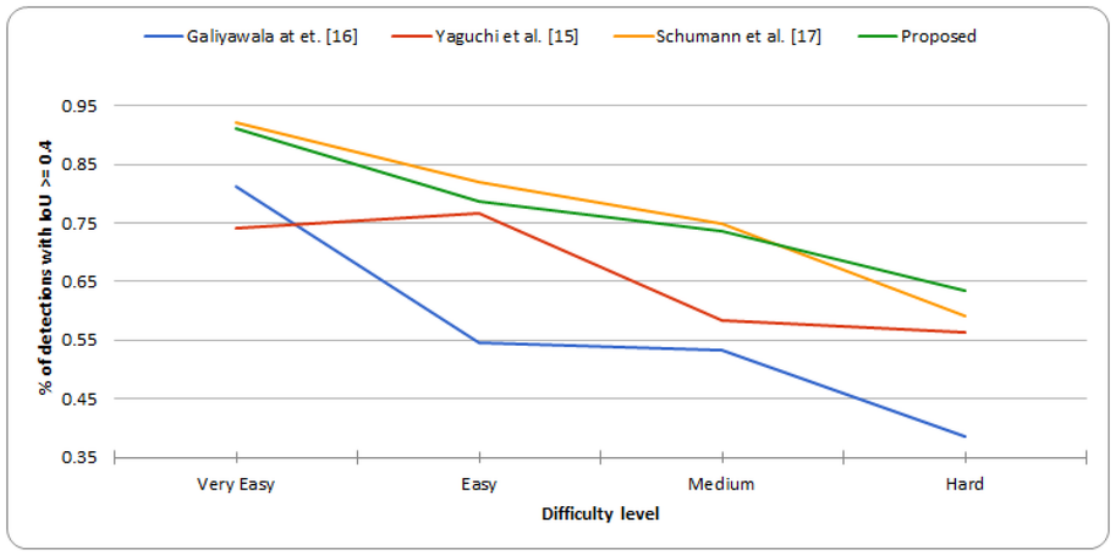

(b)

**Fig. 10.** Difficulty level wise performance comparison. (a) Average IoU and (b) % of detections with $IoU \geq 0.4$.

It uses the bounding box of the previous frame to predict box in the current frame. The box detected in the previous frame and Mask R-CNN generated boxes in the current frame are used to compute IoU. The person with maximum IoU score is considered as the match in the current frame. The IoU based regression works effectively only when Mask R-CNN detects bounding box in the current frame and soft biometrics detects correct match at least once in the earlier frames.

### 2.4. Implementation details

This section discusses the dataset, pretrained model used for transfer learning and DenseNet-169 training for color and gender classification. The person detection and semantic segmentation are done using Mask R-CNN. It is trained on MS COCO dataset [28] and tuned to detect only person class out of available 80 classes in the dataset. Weights of the DenseNet-169 pretrained on ImageNet [18] dataset are used to fine-tune the model separately for color and gender classification. The machine with the Intel Xeon core processor and 16 GB NVIDIA Quadro P5000 GPU are used to train the model.

#### 2.4.1. Dataset overview

The proposed approach uses AVSS 2018 challenge II [5] dataset for semantic person retrieval using soft biometrics. The dataset contains two tasks: (1) person re-identification (task-1) - identify a person using the semantic description from an image gallery and (2) surveillance imagery search (task-2) - localize a person using the semantic description in a given surveillance video. This paper focuses on task-2. The surveillance videos are recorded with varying level of challenges using 6 surveillance cameras. These cameras are calibrated using Tsai [24] camera calibration approach to extract height. Training dataset includes video sequences of 110 persons and the testing dataset includes 41 persons. Each training sequences con-tain annotation of 16 soft biometric attributes and 9 body markers to localize the person in the frame. The 9 body markers are: top of the head, left and right neck, left and right shoulder, left and right waist, approximate toe position of the feet. The 16 soft biometric attributes are age, gender, height, build, skin, hair, torso color, torso second color, torso type, torso pattern, leg color, leg second color, leg type, leg pattern, shoe color, and luggage. The paper uses height, torso color, torso second color, torso type, and gender to retrieve the person.

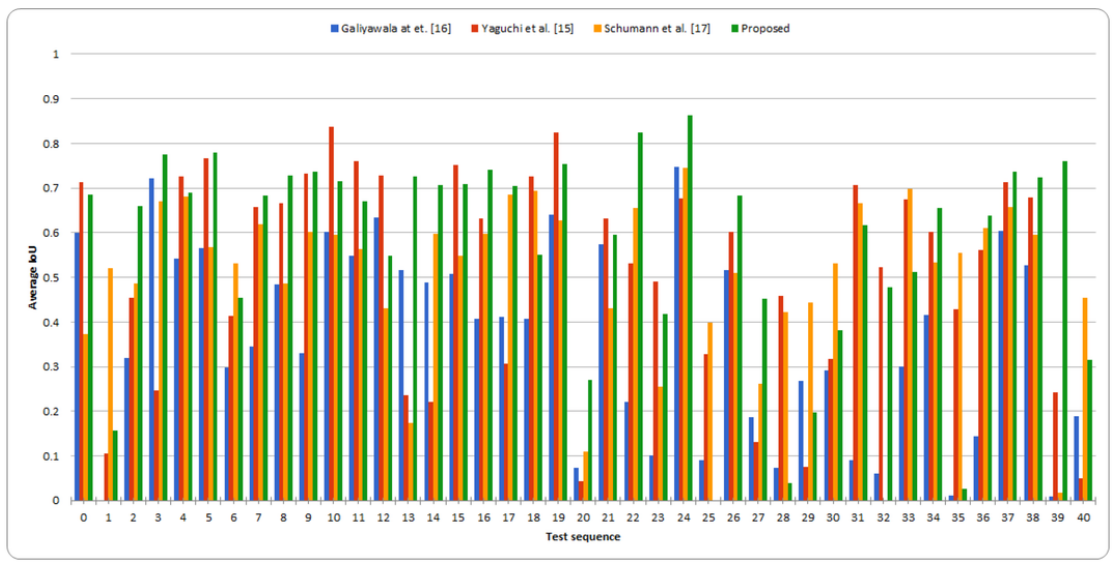

**Fig. 11.** Localization accuracy for each test sequence.

Table 2 shows classes for each soft biometric used in the proposed algorithm. Fig. 5 shows the distribution of height, torso type, torso color-1, torso color-2 and gender in the training dataset. It showcases the class imbalance for each soft biometrics. The height and torso type approximate the Gaussian distribution. It shows that samples of average height and short sleeves are more compared to other classes. There is a large variability for both torso color-1 and torso color-2. The torso color-2 is unknown for most samples. Male and female gender class also shows the minor difference in the number of samples. Test dataset sequences also have soft biometric attributes and body markers. Soft biometrics are used to create a description and body markers are used to create a ground truth bounding box during testing.

### 2.4.2. DenseNet-169 training for color classification

AVSS 2018 challenge II [5] provides 1704 color patches categorized into 12 culture colors. Additional patches are extracted using body markers from the training dataset. Total 16,369 color patches are generated. These color patches are augmented by adjusting the intensities with a Gamma factor of 0.7, 1.2 and 1.5. Data augmentation helps to avoid over-fitting during training and handles color changes due to illumination conditions. Original color patches with data augmentation generated 65,476 color patches to fine-tune the last layer (fc6) of DenseNet-169. The model is trained for 20 epochs with batch size of 16. Other network parameters are $Denseblock=4$, $growthrate=32$, $filters=64$, $reduction=0.5$, $weightdecay=0.0001$, $dropout=0.2$. Stochastic Gradient Descent (SGD) optimizer is used with $learningrate=0.001$, $decay=10^{-6}$, $momentum=0.9$, $Nesterov=True$.

### 2.4.3. DenseNet-169 training for gender classification

DenseNet-169 training is accomplished using full body images for gender classification. AVSS 2018 challenge II [5], RAP [17], PETA [18], and DukeMTMC-reID [21] datasets are used to collect 1,45,386 full body images. Each image is resized to $350\times140$ resolution to preserve the spatial ratio of full body. 80% of images are used for training and 20% for validation. The last layer (fc6) of DenseNet-169 is fine-tuned for gender model. The model is trained for 20 epochs with batch size of 16. Other network parameters are $Denseblock=4$, $growthrate=32$, $filters=64$, $reduction=0.5$, $weightdecay=0.0001$, $dropout=0.2$. Stochastic Gradient Descent (SGD) optimizer is used with $learningrate=0.001$, $decay=10^{-6}$, $momentum=0.9$, $Nesterov=True$.

## 3. Experimental results and discussion

This section discusses qualitative and quantitative results derived from the 41 test sequences. The proposed algorithm results are also compared with the AVSS 2018 challenge II participants [14-16] results. The performance of the proposed algorithm is evaluated using the True Positive Rate (TPR) and IoU. A TPR is calculated as per Eq. (1)

$$TPR(\%) = \frac{Frames\ with\ correct\ retrieval}{Total\ frames} \times 100 \qquad (1)$$

IoU is the measure of person localization accuracy where bounding box output is generated for each frame. It is a ratio of the overlapping area between the bounding box and ground truth and their union. It is measured as per Eq. (2).

$$IoU = \frac{GT \cap D}{GT \cup D} \qquad (2)$$

where, $D$ = bounding box output of the algorithm and $GT$ = ground truth bounding box.

The test sequence dataset contains varying frames numbering up to 290. The person of interest may not be present in each frame of video sequences and hence initial 30 frames are utilized for background modeling. Also, these 30 frames allow a person to completely enter in the camera field of view. Ground truth bounding box during testing is created using body markers. The y-coordinate of the head and the lowest y-coordinate of the feet construct height of the bounding box. The width is constructed using two most extreme markers from either of the neck, shoulder, waist or feet.

The results are compared using average IoU, and % detection with $IoU \geq 0.4$. The AVSS 2018 challenge II [5] divides test sequences into four difficulty levels; *very easy, easy, medium* and *hard*. The proposed algorithm is analyzed and compared with [14-16] over various difficulty levels.

### 3.1. Qualitative and quantitative results

The sample images of a correctly retrieved person using soft biometrics are shown in Fig. 6. In some test sequences, unavailability of torso color-2 is indicated as "NA" in subsequent discussions. Fig. 6 (a) shows a person from TS.19, F.36 with semantic description height (average; 160–180 cm), torso type (short sleeve), torso color-1 (white), torso color-2 (gray) and gender (male). The difficulty level is *very easy* as the target person is clearly visible, the scene has no crowd and simple environmental factors. In such cases, the algorithm retrieves the person using single soft biometric i.e., height. The TS. 39 (Fig. 6 (b)) is the *easy* level where though the scene has many persons, but the target is distinctly visible. The semantic description is height (short; 150–170 cm), torso type (short sleeve), torso color-1 (brown), torso color-2 (NA) and gender (male). The height filter output shows the multiple persons with the same height class. The person of interest is then retrieved using torso color-1. In this case, the algorithm uses two soft biometrics i.e., height and torso color for the detection.

A person in the TS.7, F.95 with semantic description height (short; 150–170 cm), torso type (long sleeve), torso color-1 (gray), torso color-2 (white) and gender (female) is in Fig. 6 (c). The scene is with *medium* difficulty as crowd flow is fair and multiple persons matching the description are present. The algorithm retrieves the person using height and torso color-2.

Fig. 6 (d) shows a person from TS.22 F.30 with semantic description height (average; 160–180 cm), torso type (short sleeve), torso color-1 (yellow), torso color-2 (black) and gender (male). The difficulty level is *medium*. Color filter output produces couple of matches. Such cases are refined using a gender filter. Fig. 6 (e) shows the sample for *hard* difficulty level where crowd flow is medium, persons with partial match are present and person of interest is occluded. Fig. 6 (e) shows a person from TS.20 F.50 with semantic description height (short; 150–170 cm), torso type (short sleeve), torso color-1 (white), torso color-2 (NA) and gender (female). The height filter outputs 3 matches, color filter produces 2 matches and finally the person of interest is retrieved using gender filter. Fig. 6 (d) and (e) depicts the utilization of all soft biometrics i.e., height, torso color, and gender to retrieve the person.

Fig. 6 (g) depicts the person retrieval using IoU based box regression. It shows a person from TS.11 F.85 with semantic description height (short; 150–170 cm), torso type (short sleeve), torso color-1 (blue), torso color-2 (NA) and gender (female). The height filter outputs 2 persons. Here, the color filter fails to retrieve the person. Thus, IoU based box regression calculates the IoU score between bounding box with match detected in the previous frame (i.e. F.84, Fig. 6 (f) and Mask R-CNN boxes detected in the current frame (i.e., F.85, Fig. 6 (g)). A total of 5 boxes are generated by Mask R-CNN and their IoU scores are 0, 0, 0.9334, 0, and 0. The box with maximum score 0.9334 is considered as correct retrieval as shown in Fig. 6 (g).

Fig. 7 shows the sample frame of sequences where the algorithm fails to retrieve the person of interest due to various challenges like low illumination, merging of person with the background, occlusion and heavy crowd flow. The person of interest is with a red bounding box in each frame.

### 3.2. Performance analysis

The TPR (%) for all test sequences is shown in Fig. 8. The algorithm achieves average TPR of 76.21% which is 40.81% higher compared to average TPR 54.12% of [15]. Among 41 persons, 32 persons are retrieved with TPR greater than 60%, 5 with TPR between 30% and 60% and 4 with TPR between 0% and 30%. It can be observed that for TS.1, 25, 28, 35 and 40; the algorithm could not achieve good performance due to challenges shown in Fig. 7.

The proposed approach and the AVSS 2018 challenge II approaches [7, 14-16] are compared to evaluate the performance. Table 3 compares average IoU and $IoU \geq 0.4$ as an evaluation metrics. The proposed approach outperforms all previous approaches in terms of average IoU. It achieves 0.569 average IoU, which is better than state-of-the-art average IoU of [14]. The algorithm also achieves similar performance in terms of detection with $IoU \geq 0.4$ when compared to state-of-the art approach of [16]. One must note that the approaches in [14, 16] use full set of soft biometrics while the proposed approach uses only 4 attributes (i.e., height, torso type, color, and gender) to achieve state-of-the-art performance. It also reduces the hardware pipeline required for implementation of the algorithm.

Fig. 9 shows the difficulty level wise distribution of the test sequences. The medium and hard level cover close to 51% of test sequences. It means that the developed algorithm should be reasonably robust to achieve good classification accuracy. The algorithm performance is also compared (Fig. 10) with respect to difficulty level. It can be observed from Fig. 10a that the proposed approach outperforms all previous approaches [14-16] for all difficulty levels in terms of average IoU. The performance increases by 14.69%, 1.32%, 13.12% and 10.35% for difficulty levels; *very easy, easy, medium, and hard* respectively over the state-of-the-art methods [14-16].

The adaptive torso patch extraction, torso patch augmentation, and IoU based box regression are the key reasons for the achievement of better average IoU. It can be observed from Fig. 10b that the algorithm also achieves similar performance in terms of detection with $IoU \geq 0.4$ when compared to state-of-the-art approach of [16]. The use of tracking in [16] helped to achieve better performance for *very easy, easy* and *medium* level while the proposed approach outperforms [16] for the *hard* level. The *hard* level has sequences with multiple matches (i.e., partial match to description cf. Fig. 6 (e)). The tracker in [16] generates noisy bounding boxes, while IoU based box regression helps to predict the correct bounding box in the proposed approach.

The localization accuracy is compared in Fig. 11 for each test sequences. The proposed algorithm achieves highest average IoU for 20 sequences {TS.2, 3, 5, 7, 8, 9, 13, 14, 16, 17, 20, 22, 24, 26, 27, 34, 36, 37, 38, 39} out of 41. The above sequences cover all the difficulty levels in the dataset. The color classification of the proposed algorithm works well for these sequences. It achieves the increase of 0.67% to 214.1% in average IoU when compared to the best IoU from [14-16]. The approach shows the exceptional performance for TS.2, 13, 27 and 39 compared to previous approaches [14-16].

As discussed in Section 2.4.2, the color model is trained with data augmentation to handle the color intensities changes in the surveil-lance. Thus, it gives better results for TS. 27 and 39. The TS.2 has low illumination and merging between foreground and background. The proposed approach gives good results over TS.2 due to better color model and bounding box regression. The adaptive torso patch extraction helps to achieve better performance for TS. 13 and 16. The torso type is "no sleeve" for these sequences.

The algorithm fails to achieve good performance for TS. 1, 25, 28, 35 (Fig. 7). TS.1 in Fig. 7 (a) shows challenges of low illumination, occlusion, and person merging with background. In such cases, detection fails or it generates a single bounding box covering occluded persons. Fig. 7 (b) (TS.25) and (c) (TS.28) also depicts cases of occlusion. The TS.35 shown in Fig. 7 (d) shows dense crowding. Fig. 11 showcases performance variation for each test sequence. The specific approach may turn out to be the best candidate solution for the sequence.

Test sequences TS.20, 23, 25, 27, 28, 29, 40 pose a challenge for all approaches [14-16] and also the proposed approach. The average IoU is less than 0.5 for these sequences. The maximum average IoU is 0.27 [proposed approach], 0.49 [14], 0.39 [16], 0.45 [proposed approach], 0.45 [14], 0.44 [16] and 0.45 [16] for TS.20, 23, 25, 27, 28, 29, 40 respectively.

## 4. Conclusion

This paper introduces adaptive torso patch extraction and IoU based bounding box regression for person retrieval with the fewer traits. The proposed approach performs better at all difficulty levels and achieves best average IoU of 0.569 which is 11.35% higher over the state-of-the-art average IoU 0.511 of [15]. It also achieves competitive performance in terms of detection with $IoU \geq 0.4$ when compared to state-of-the-art approach of [16]. Due to linear filtering approach, an error from one filter will propagate to further stages. Also, an incorrect retrieval in the previous frame will propagate to the current frame and so is the case with IoU based bounding box regression. Future work can focus on person retrieval with challenges like occlusion, color variations, varying illumination conditions, persons with similar appearance and dense crowding. It may incorporate more soft biometrics (e.g., leg color) for person retrieval. Softer decision by weighing each soft biometric appropriately can be applied in cascaded filtered approach. The limitation of tracking and IoU based regression can be removed by fusing final decisions.

**Acknowledgments**

This work is supported by the Board of Research in Nuclear Sciences (BRNS) of India (36(3)/14/20/2016-BRNS/36020). The authors acknowledge the support of NVIDIA Corporation for a donation of the Quadro K5200 GPU used for this research. We would also like to thank the AVSS 2018 challenge II organizers for creating the challenging dataset and providing the necessary details to compare the performance.